\documentclass[12pt]{article}
\usepackage[utf8]{inputenc}
\usepackage{amsmath}
\usepackage{graphicx}
\usepackage{hyperref}
\usepackage{natbib}
\usepackage{geometry}
\usepackage{setspace}
\usepackage{authblk}
\usepackage[T1]{fontenc}

\title{On the stochastics of human and artificial creativity}

\author[1]{Solve Sæbø}
\author[2]{Helge Brovold}

\affil[1]{KBM, Norwegian University of Life Sciences, Ås, Norway}
\affil[2]{Cqpsych AS, Trondheim, Norway}

\begin{document}

\maketitle

\begin{abstract}
What constitutes human creativity, and is it possible for computers to exhibit genuine creativity? We argue that achieving human-level intelligence in computers, or so-called Artificial General Intelligence, necessitates attaining also human-level creativity. We contribute to this discussion by developing a statistical representation of human creativity, incorporating prior insights from stochastic theory, psychology, philosophy, neuroscience, and chaos theory. This highlights the stochastic nature of the human creative process, which includes both a bias guided, random proposal step, and an evaluation step depending on a flexible or transformable bias structure. The acquired representation of human creativity is subsequently used
to assess the creativity levels of various contemporary AI systems. Our analysis includes modern AI algorithms such as reinforcement learning, diffusion models, and large language models, addressing to what extent they measure up to human
level creativity. We conclude that these technologies currently lack the capability for autonomous creative action at a human level. 
\end{abstract}

\section{Introduction}

Currently, there is an active and ongoing debate regarding whether artificial intelligence (AI) systems will achieve so-called AGI (Artificial General Intelligence) in the near future. However, this debate lacks a solid foundation, as there is no consensus on the criteria for evaluating AGI. What exactly should the 'generality' in AGI systems encompass? Often, in discussions about AGI, comparisons to 'human-level' performance are frequently made. Thus, it becomes crucial to explore what constitutes human intelligence. Perhaps understanding AGI is more feasible when juxtaposed with conventional interpretations and definitions of human intelligence.

Moreover, opinions on the nature of human intelligence vary. According to Encyclopedia Britannica, human intelligence is described as:

\vspace{5mm}

\noindent\textit{Human intelligence is, generally speaking, the mental quality that consists of the abilities to learn from experience, adapt to new situations, understand and handle abstract concepts, and use knowledge to control an environment.}

\vspace{5mm}

In the literature, multiple theories and conceptualizations of human intelligence exist. Prominent ones, such as those proposed by Spearman (1923) and Sternberg (1985), as well as later theories reflected in the definition provided by Britannica, commonly include the ability to adapt to the environment, generalize to new situations, and problem-solving as integral components of intelligence. For example, Sternberg's Triarchic Theory of Intelligence encompasses these aspects within the experiential or creative subtheory of intelligence.

Without delving deeply into this extensive field, which encompasses cognitive, psychometric, philosophical, and biological perspectives,we will simply assert that creativity appears to be an appreciated component of human intelligence. Building on this premise, we argue that for any artificial system to achieve human-level performance, or AGI, it must also attain a comparable level of creativity.

As the reader might be aware, the term 'Artificial Intelligence' (AI) traces its origins back to 1955, when mathematician John McCarthy and colleagues proposed a research project at Dartmouth College for the summer of 1956. The funding proposal for this summer school stated:

\vspace{5mm}

\noindent\textit{The study is to proceed on the basis of the conjecture that every aspect of learning or any other feature of intelligence can in principle be so precisely described that a machine can be made to simulate it.}

\vspace{5mm}

This ambitious conjecture encompassed various facets of learning and intelligence, such as language processing, neural networks, and abstraction, which continue to be pivotal in contemporary AI research. Perhaps less widely recognized is the fact that the Dartmouth proposal also delved into the realm of 'Artificial Creativity' (AC): 

\vspace{5mm}

\noindent \textit{A fairly attractive and yet clearly incomplete conjecture is that the difference between creative thinking and unimaginative competent thinking lies in the injection of some randomness. The randomness must be guided by intuition to be efficient. In other words, the educated guess or the hunch include controlled randomness in otherwise orderly thinking.}

\vspace{5mm}

Thus, the proposal posited that creative problem-solving, a defining feature of human intelligence, might involve an element of randomness, steered by intuition – a conjecture that alludes to the intricate balance of order and chaos in the creative process.

Evidently, McCarthy and his team had their own inklings about addressing both AI and AC, ambitiously planning to unravel these challenges in a relatively short span, aiming to develop a computer program by the end of the two-month seminar in 1956. However, this task proved to be more challenging than anticipated. Now, nearly seven decades later, we are compelled to ask: How close have we come to achieving McCarthy’s vision of creating AC? Are we on the brink of this breakthrough, or is it still a distant goal?

Traditionally, creativity has been viewed as a realm exclusively belonging to human intelligence and artistic endeavor. Yet, the emergence of sophisticated AI technologies, capable of generating art, music, and literature, is increasingly blurring the distinction between human and artificial creativity. This ambiguity is further amplified by mainstream and social media, which often depict generative AI as 'creative.' Also the research community occasionally add to this confusion, a recent example being Hubert et al. (2024) with a title stating that generative language models have become more creative than humans on divergent thinking tests. Confusion with regard to the extent AI has become creative is obviously partly due to differing semantics. What do people really mean when they express that some image made by generative AI is creative? And a closer look into Huber et al. (2024) reveals that the title of the paper actually points to AI having higher 'creative potential' than humans on certain tasks, which is a term that better reflects the results of their interesting study. 

Obviously, to effectively compare human and artificial creativity, it is imperative to establish a more precise definition of 'creativity'. One important distinction is between creativity in the \textit{loose} and the \textit{strict} sense. The latter is perhaps best defined by the difference to creativity in the loose sense, which refers to activities that society has acknowledged as 'creative' in their nature. It is accepted that making fine art, poetry, music are creative activities (Ritchie, 2019), whereas doing, for instance, research is not. However, we believe most readers would agree that creativity is indeed required in doing cutting edge research. In the loose sense AI may be regarded as creative in the way it can generate artwork, write poems or make music. However, to what extent computer systems can exhibit creativity in the strict sense has been a field of active research, at least since the seminal work of Boden (2004). This has given rise to its own discipline called Computational Creativity (CC) which covers research on both strict CC as well as human-computer relations in creative processes. Although CC is an established term within this discipline, we will use artificial creativity in this paper (AC) as an extension of AI and as part of AGI. The work of Boden and others within CC is still highly relevant.

The primary aim of this paper is to first establish an understanding of human creativity in the strict sense. This endeavor itself covers a vast research area, encompassing disciplines such as philosophy, psychology, biology, and neuroscience. In our pursuit, we seek to comprehend human creativity in a way that lays the groundwork for a subsequent examination of artificial creativity. To achieve this, our goal is to construct an algorithmic understanding of the human creative process, viewed through a statistical lens. More precisely, we intend quite heuristically to conceptualize human creativity as a stochastic process within a Bayesian inference framework. In doing so, we will draw upon knowledge and methodologies from not just statistics, but also psychology, philosophy, neuroscience and chaos theory. Armed with this framework, we will then proceed to evaluate the latest advancements in AI, particularly those related to generating novel content, and determine whether the generative processes and their outcomes can be classified as 'creative'. 

The motivation for understanding human creativity is thus twofold. First, it allows us to establish a baseline against which AI's creative capabilities can be measured. Second, it provides philosophical insights into the nature of creativity itself, which is crucial for evaluating the advancements in AI. As we continue to push the boundaries of what AI can achieve, revisiting the foundational ideas of the Dartmouth seminar and exploring the depths of human creativity becomes imperative for a holistic understanding of AC.

\section{On human Creativity}

The 'founders' of AI, John McCarthy and colleagues, characterized creativity as involving '...randomness... guided by intuition...' in their proposal for the Dartmouth summer seminar on Artificial Intelligence in 1956. Their hypothesis suggested that the creative process entailed a form of controlled randomness. This description encapsulates two critical and contrasting facets of creativity, which we will revisit in this exploration of AC.

Many artists, writers, composers, scientists, and others have experienced the unpredictable nature of creativity, where inspiration is fleeting and elusive. Despite its enigmatic character, it's often observed that certain conditions or external factors can catalyze creative thoughts. Consider, for instance, Pablo Picasso, who drew inspiration from his muse, Marie-Thérèse Walter, or Albert Einstein, who found that leisurely walks in the campus park helped his thoughts to wander, often leading to novel ideas upon his return to the office. Therefore, it seems possible to stimulate, nurture, or even guide creativity, despite its inherent randomness, although it remains somewhat mystical, appearing and disappearing unpredictably. Even more mysterious is the 'eureka moment' – a sudden, profound insight, akin to Archimedes' legendary moment of realization in the bathtub. This is the instant of clarity that flashes before the mind’s eye, appearing seemingly from nowhere, presenting a vivid solution to a previously intractable problem.

Numerous studies have been conducted to unravel the secrets of creativity, elucidating the conditions under which it manifests and determining whether individuals possess varying potentials for creativity. Various assessments, such as the Torrance Tests of Creative Thinking, have been developed to measure attributes believed to correlate with creativity, including the level of divergent thinking, or more commonly, the ability to 'think outside the box'.

Researchers in the fields of modern neuroscience and psychology have utilized brain imaging techniques such as EEG and fMRI to shed light on the processes underlying creativity and the eureka moment, as well as identifying the brain regions and neural networks involved. However, a comprehensive understanding of the complex cognitive processes behind creativity in general, and the mysterious eureka moment in particular, is still lacking.

Could statistical theory provide a pathway to deciphering some of the mysteries of creativity? The seemingly spontaneous emergence of creative thought is likely the outcome of stochastic (random) cognitive processes within the brain, processes that are extensively examined within the realm of statistics as part of stochastic theory. Both conscious and unconscious associative processes seem to lie at the heart of creativity broadly, including the eureka moment. The well-documented characteristics of stochastic processes and statistical theory might offer valuable insights into the mechanisms of human creativity, particularly when integrated with findings from neuroscience, psychology, and philosophy.

Turning to the philosophical perspective, considerable discourse surrounds the essence of creativity. However, the contemplations of physicist and philosopher David Bohm stand out as among the most profound insights on the subject. His perspectives are gradually gaining acknowledgment in contemporary neuroscience.

\subsection{A philosophical view on human creativity}

As a scientist Bohm was well known with the scientific methods, and to him artistic creativity and scientific progress were tightly connected. Just as science typically proceeds through the two stages of 1) hypothesis generation, and 2) deductive reasoning and experimental testing, Bohm argued that a creative process also contains two main stages, and he referred to them as 1) insight, and 2) fancy. Bohm divided each of these further into two subtypes, the imaginative and the rational insight or fancy.

The quote by Spanish painter Joan Miró, \textit{'The works must be conceived with fire in the soul but executed with clinical coolness'}, aptly encapsulates these dual aspects of the creative process as interpreted by Bohm. Our interpretation of Bohm is that the first stage, insight, describes how novel ideas or insights develop through an open process of divergent thinking, either consciously (rational insight) or unconsciously (imaginative insight), with the latter often manifesting as what is known as a ‘Eureka moment’. The second stage, fancy, describes a more deductive process, where insights are explored and hammered out into matured ideas, innovative artefacts or theories, either by semi-conscious unfolding (imaginative fancy) or through fully conscious and focused reasoning, prediction and evaluation (rational fancy). As Einstein experienced an imaginative insight, visualizing a unifying order that could integrate disparate theoretical fragments, a new creative phase was initiated. This phase involved deducing the broader implications of his insight for physics, a task that demanded concentrated effort and collaboration. 

The stages of insight and fancy described by Bohm delineate various phases in the creative process but do not explicitly define creativity \textit{per se}. What distinguishes a creative thought from any other thought? In psychology, a common interpretation of creativity is that \textit{creativity is novelty that works}. This means that a creative act or idea must not only present a certain degree of novelty but also not everything 'new' is inherently creative. It's a frequent misunderstanding to equate randomness with creativity. There needs to be an additional element. It must somehow 'work'. In science, it should pass empirical tests and contribute to the expansion of knowledge within a research field; in art, it should evoke an engaging response from viewers; and in music, it should offer listeners chills, a profound sense of harmony, or even the urge to dance. Thus, we can assert that creativity involves exploring the unknown to discover something new and beneficial. A 'novelty that works' ought to aid in broadening our perspectives. 

Human creativity can be envisioned as a two-phase process. The first phase is characterized by an open, divergent exploration into the unknown, often happening in a semi-conscious or unconscious state. This stage generates hypotheses about new structures or patterns, boldly venturing into unexplored territories without predetermined limits—bounded only by the scope of one’s imagination. The subsequent phase is evaluative, concentrating on the feasibility of the newly generated idea. Critical questions like “What does it imply?”, “Does it work?”, and “Is it beneficial?” become central. If the idea proves to be practical, it might be embraced in a way similar to the scientific method, where a hypothesis is maintained as long as its predictions concur with empirical evidence. Joan Miró aptly described this stage as requiring “clinical coolness,” a focused and incisive mindset to polish and scrutinize the new idea or hypothesis.

Typically, incremental innovations are more readily embraced than radical shifts. Minor modifications to existing scientific theories are more likely to survive the scrutiny of peer review than propositions advocating for a new paradigm. Yet, it is often the more audacious, paradigm-shifting ideas that are retrospectively recognized as marks of true genius and creativity. This suggests the existence of different levels or types of creativity, a subject we will revisit later.

David Bohm also describes creativity as a process that brings order and structure into chaos, like Einstein who introduced his new order in physics with his theories of relativity. Throughout our lives, we learn to navigate the randomness and variability in our environment, yet the creative discovery of order or structure diverges significantly from the conventional learning experiences in educational settings, which are typically characterized by \textit{supervised learning}, where a teacher organizes and structures knowledge for us.  On the contrary, creativity is rather an example of \textit{unsupervised} learning. In statistics unsupervised learning means that a ‘truth’, often called the gold standard,  is unknown. In the absence of a gold standard, chaos must be boldly explored and new order must be discovered, by ourselves. 

Nonetheless, without any guidance with regard to the nature of the unknown order, the search is likely futile. Some assumptions must be made about the properties of the unknown. Just as unsupervised learning in statistics relies on assumptions like the probability distribution of data, the level of entropy (order and disorder) or the number of unknown categories to be found. John McCarthy pointed at intuition as one guide for creativity, although this leaves us the open question of ‘what is intuition?’ David Bohm on his side states that scientists, and artists alike, seek a sense of wholeness, harmony or beauty in their work:

\vspace{5mm}

\noindent \textit{[…about creativity]…, I suggest that there is a perception of a new basic order that is potentially significant in a broad and rich field. The new order leads eventually to the creation of new structures having the qualities of harmony and totality, and therefore the feeling of beauty.}

\vspace{5mm}

Both Bohm and McCarthy point to similar control mechanisms that are involved in order to limit the stochasticity of the divergent thought during the stages of the creative process, namely our personal or common notions of what feels right, what is normal, what is within ethical standards, or relevant and interesting, or, say, according to common sense or being commonly acceptable. Our control functions are our biases, inherited or learned, narrow or wide, conscious or unconscious. These are our learned abstractions and our internal models of how the world is and how it will evolve. It is the hierarchical structures of biases we use to deem some new idea as useful or not.  They represent our frames of reference, helping us to bring order into chaos, to survive…., and to be creative. This is our 'World model'.

In summary, creativity can be described as a dynamic interplay between order and chaos, a balancing act between the quest for novelty and the assessment of utility, and a navigation between randomness and regulation. Bohm elucidates that the creative process is an ongoing iterative movement between stages of insight and fancy. To integrate these concepts, each phase—whether it involves imaginative or rational insight, or imaginative or rational fancy — features varying degrees of stochasticity and control. Imaginative or rational insight is marked by high randomness and low control, in contrast to imaginative or rational fancy, where the balance shifts towards greater control and lower stochasticity.

\subsection{A statistical view on human creativity}

\subsubsection{The stochastics of divergent thinking}
\label{divergent}

Creativity is seemingly a process that resides along the edge between chaos and control. In this section we will look closer into the stochastic, or chaotic, part of this process.  Divergent thinking or 'thinking outside the box' is often used to describe this somewhat unpredictable exploration of the unknown in the pursuit of potential new, creative ideas. To delve deeper into the stochastic component of human creativity, exploring the field of stochastic processes within statistics could prove insightful. This exploration is particularly pertinent if we aim to build upon the conjecture posited by McCarthy and his colleagues in 1956, suggesting that human creativity could, to some extent, be emulated by computers. Investigating stochastic processes can help us gauge the extent to which such computational imitations need to mirror the biological or algorithmic intricacies of the human creative process. Understanding these parallels and distinctions is crucial for advancing our comprehension of how closely artificial systems can replicate the dynamism and complexity of human creativity.

Let's begin with a straightforward example of 'novelty that works' from statistics: the generation of random numbers on a computer. Statistical software can generate numbers as a random sample from a probability distribution, such as a normal distribution. Various methods have been developed for random sampling, and one prominent category is the Markov Chain Monte Carlo methods (MCMC) (e.g. Gilks et al., 1995). These methods are widely used in Bayesian modeling for statistical inference. They involve computer-intensive procedures where a so-called random walk process is initiated to simulate a dependent chain of observations from a specific probability distribution, often referred to as a target distribution. A frequently utilized algorithm for this purpose is the Metropolis-Hastings algorithm (Hastings, 1970).

In the typical configuration of MCMC, each successive step of the Markov chain is influenced by the previous one through a current value, which is either a starting point or the most recently sampled value. Given this current value, the Metropolis-Hastings algorithm usually alternates between two steps:

\begin{enumerate}
    \item A new candidate value, typically depending on the current, is drawn from a proposal distribution (a random step).
    \item The candidate is accepted or rejected as a new value of the chain in light of the target distribution (an evaluation step).
\end{enumerate}

This Markov chain setup mirrors an associative chain of thoughts, where the proposal distribution chiefly governs the randomness in the creative process, and the target distribution (or what we might term the bias) assesses usefulness. This provides a statistical framework for establishing association processes that fulfill both criteria of a creative process as previously discussed. What is essential is a sensible method for generating candidate ideas and a mechanism to evaluate the usefulness of these ideas.

The cognitive parallel to MCMC could be seen as 'thinking inside the box'. It resembles a scenario where we sit and ponder, creating a chain of associations where all thoughts are judged against the same criterion of usefulness, which represents our current personal cognitive bias. Typically, we describe creativity as 'thinking outside the box'. However, as we will explore further, thinking outside the box essentially involves challenging our existing frameworks—by transforming the biases we apply to evaluate our thoughts. We will revisit the significance of biases later. For the moment, let's leverage the MCMC example and the statistical framework of Markov models to delve deeper into the stochastic nature of creativity.

Introspective thinking, characterized by limited external stimuli, is a fully associative process where one thought naturally leads to another. To recall memories, we cannot simply look them up as we would with the index of a dictionary; instead, we must access them through associations. In neuroscience, it is commonly understood that thoughts and memories are stored in long-term memory as pathways of strongly connected neurons, sometimes referred to as \textit{engrams}, organized within a vast network of associations.

Donald Hebb, the Candadian neuro-psychologist, articulated the process of learning at the neuronal level in 1949 with the principle: 'neurons that fire together, wire together'. The Hebbian theory explains how synapses, the connections between neurons, are strengthened during the learning process when two connected neurons are activated simultaneously or in succession during experiences or learning activities. Conversely, the reversed phrase 'neurons that wire together, fire together' might more accurately describe the replay processes that occur when we recall memories and learned facts. Random signal processes tend to follow well-established routes, manifesting either as a conscious chain of thoughts or within the realm of our unconscious. Notably, chains of connected memories are also replayed during sleep, as demonstrated in studies with rodents by Wilson and McNaughton (1994).

The likelihood of transitioning from the current thought to other thoughts can be described as transition probabilities in statistical terminology. Initially, let's presume that any thought at a given moment is selected from a fixed array of potential thoughts and memories, which we can term the state space of thought. The probability of various cognitive outcomes from this state space is influenced by an individual's entire history of experiences, collectively shaping the distribution of transition probabilities across the personal state space of thought. For simplicity, we initially assume that these transition probabilities remain constant over time. While some transitions may be highly probable, others might be quite unlikely. Based on this distribution, a random decision is made regarding the next thought to consider.

As indicated by our brief statistical simulation analogy above, Markov processes can, to a certain degree, mimic cognitive processes. Markov processes possess numerous intriguing properties that are also cognitively relevant to creativity. Here, we will outline some of these properties.

\vspace{5mm}
\noindent\textbf{The primer effect} 

\noindent The first property is the ‘primer effect’. A random walk process must start somewhere, hence it requires a starting value, or a so-called primer. From the primer the process ‘walks’ from one value to another. Accordingly, for a cognitive process the current thought is very often the primer for the next thought. 

The priming effect plays a significant role in the realm of human creativity. Without external stimuli, thought chains tend to be introspective, predominantly navigating through well-trodden areas within our state space of thought. In this state, we are not inclined towards novelty. However, larger leaps in creativity may be facilitated if we engage with our surroundings or actively seek out new experiences, such as by exploring new cultures through travel, choosing a different route to work, or selecting a random book from the library. Such actions can initiate thought chains that venture into less explored areas of our state space of thought. Kounios and Beeman explore various factors that can enhance insightfulness in their book 'The Eureka Factor: Aha Moments, Creative Insight, and the Brain' (Kounios and Beeman, 2015).

\vspace{5mm}
\noindent \textbf{Focus level} 

\noindent Another crucial aspect of stochastic processes is the step length of the walk, typically dictated by the proposal distribution and subsequently constrained by the target distribution. This can be considered the focus parameter of the process. When step lengths are short, the process traverses the state space slowly, engaging only with closely connected states over extended periods. As a result, the sequence of visited states exhibits a high level of auto-correlation, meaning in a cognitive context, thoughts tend to be similar and related over time. A person with highly auto-correlated thinking could be described as narrow-minded. However, being narrow-minded might be essential, for example, when we need to concentrate intensely on solving a challenging problem or focus on learning a new skill.

Neurologically, strong focus is achieved in the brain by the activation of inhibitory neurons through the increased release of the neurotransmitter GABA (Gamma-aminobutyric acid), which lowers the transition probabilities for long step transitions to thoughts that appear irrelevant or distracting. GABA effectively blocks out external and distracting signals, particularly during stages of fancy, as discussed by David Bohm, especially during what he terms rational fancy. If step lengths are permitted to increase (due to a decrease in GABA), focus diminishes, and a more diffuse state of mind is induced, as in the stages of insight.

The issue with the slowly progressing cognitive chain of a focused mind lies in the high likelihood of overlooking creative potential solutions to problems. This is because it takes too much time to navigate the relevant sections of the thought state space. Conversely, excessively long steps might elevate the chance of generating very distant ideas, which could be dismissed as irrelevant in the given context (i.e., an unlikely proposal according to the target distribution). 

\vspace{5mm}
\noindent \textbf{Cognitive flexibility} 

\noindent In Bayesian inference so-called well-mixing Markov processes are desirable, wherein the chain moves across the state space with optimal step lengths, avoiding both being too narrow-minded and too diffuse. Such well-mixing processes have the largest probability of covering a relevant state space in sufficient proportions within a limited time span. 

A unique type of Markov chain is the so-called time-inhomogeneous chain, where transition probabilities may change over time. This describes association processes that alternate between focused and defocused modes. The ability to flexibly switch between short and long transitions can be beneficial for the creative process, leading to well-mixed association chains. David Bohm, in his book 'On Creativity,' characterizes the creative process as an iterative cycle between stages of insight and stages of fancy, essentially depicting a time-inhomogeneous stochastic process of thinking. This flexibility is also a crucial aspect in a popular approach to design and innovation processes known as design thinking.

\vspace{5mm}
\noindent \textbf{Plasticity and learning} 

\noindent For cognitive flexibility we relaxed the initial assumption that there is a fixed probability distribution across an individual’s state space of thought. Let’s now also relax the assumption that this state space of thought itself remains static. These initial assumptions were characteristic for what is known as a stationary distribution in stochastic process theory. However, it is unrealistic to consider the cognitive state space as static or the thought distribution as stationary. This is because we continually expand and modify our state space through learning, and the brain is constantly evolving, both in its functions and structure. The associative chains of thought actually alter the probability distribution over the state space as they progress. This is because repetitive engagement in chains of association plays a crucial role in Hebbian learning, where the process of both learning and recall change the synaptic strengths between neurons. Simply by visiting certain thoughts or memories, the likelihood of revisiting them increases.

Moreover, new and previously unexplored thoughts emerge during the random walk as a result of creative thinking or learning from external inputs. Conversely, certain parts of the state space may be nearly or completely eliminated (forgotten) due to the loss of synaptic connections, possibly from being seldom revisited or due to brain lesions. The brain's significant plasticity and capacity for change mean that the thought state space is equally dynamic. Therefore, the random walk of learning and creative thinking can be viewed as a non-stationary stochastic process. Reflecting on this, it seems evident. Over our lifetimes, our interests, values, and the contexts we are part of evolve, undoubtedly influencing our thought processes. It is reasonable to surmise that divergent thinking, which involves seeking new perspectives and expanding one's knowledge base, enhances creativity due to the inhomogeneity and non-stationarity characteristics of the associative thinking process.

\vspace{5mm}
\noindent \textbf{Inter-personal dialogue} 

\noindent Multiple Markov chains running in parallel may cover a state space faster, and integrating the information from many chains may be beneficial for many purposes in a statistical setting. Parallel cognitive processes are also integrated in the context of a conversation, for instance, at the lunch table, by the coffee machine, or during group based learning. Different people have different experiences and knowledge levels, fields of interests, values and personalities. Each person even has his or her individual cognitive state space and thought transition distribution. Through dialogue, parallel random walks of thought may evolve jointly towards a better understanding of some subject to be learned. 

However, in a dialogue, the thought processes themselves are actually not interacting, because the participants do not observe one another’s thought processes. A special kind of Markov processes, called Hidden Markov Models (HMMs), can give insight into the dialogue in that respect. In statistical inference HMMs are used to model processes where it is reasonable to assume that there is an underlying, hidden process that occasionally gives rise to observable output. Our cognitively relevant example of a HMM is one person's indirect observation of the thought process of another person. For instance, your conscious thoughts are available to you, but only occasionally and approximately observable to others, and that is whenever you orally express your thoughts. 

In the brain the process of transferring conscious thoughts from the prefrontal cortex to the cortical areas responsible for speech (Broca’s area), is itself a stochastic process that adds noise to the output. This manifests itself in the fact that sometimes it is difficult to express exactly what you are thinking. HMM’s are similarly defined not only by transition probabilities for the hidden state space, but also state-dependent probabilities for generating observable and noisy output. Hence, some thoughts are more likely to be expressed than others, and with higher or lower accuracy. 

Furthermore, it is reasonable to assume that the output probabilities are inhomogeneous, meaning that various contextual factors and conditions have an influence on the probability that thoughts are articulated. The output probabilities may, for instance,  depend on the context of the conversation, or the personalities of the participants of a dialogue. 

Creative team processes, involving dialogue and conversations, may be explored further in the context of parallel HMMs, but a main point in this context is that through conversation, associations are exchanged, which may lead to jumps in the thought processes of the participants. These jumps can result in a better coverage of the state space and faster idea generation and learning for each individual group member. 

\vspace{5mm}
\noindent \textbf{Intra-personal dialogue} 

\noindent Parallel associative processes also occur in another cognitive domain, namely in the unconscious mind. The brain is never at rest. Even when we have dreamless sleep, the neurons fire and pass signals to each other. The aforementioned replay of thoughts along probable neuronal paths, run in our unconsciousness, visiting thoughts, memories and ideas, and maybe they are related to some unresolved issues or problems you have been focusing on lately. This resembles divergent conscious thinking, but recent studies indicate that the unconscious is much more powerful than the conscious mind as a provider of novel thoughts and ideas. There are several properties that make the unconscious so effective for divergent thinking, for instance:

\vspace{5mm}
\noindent \textit{Parallel processes} 

\noindent As discussed above, parallel processes may cover a state space of thought faster than a single process. Since the unconscious mind is not restricted to give attention to a single thought process, like the conscious mind is, it is reasonable to believe that there are multiple parallel cognitive processes running beneath the surface of awareness. The psychologist and Nobel laureate David Kahneman (2011) gives support to this in his book 'Thinking Fast and Slow' where he describes the fast thinking, automated, and unconscious ‘system 1’ in the brain as a parallel processing system. The slow thinking and conscious ‘system 2’, he states, is a serial system. This multivariate thinking process of the unconscious mind may be highly beneficial if it is coupled with some information integration system. We will return to the latter aspect later.

\vspace{5mm}
\noindent \textit{Reduced bias} 

\noindent As we also will return to below, our biases may directly influence on the random walk process of divergent, conscious thinking, but it is reasonable to believe that biases are less influential on unconscious thoughts. Most of us have probably experienced divergent thinking during dream sleep which is far beyond what we perhaps would allow ourselves when we are awake. The famous psychiatrist and psychologist, Milton Erickson, who specialized in medical hypnosis, said that biases are the province of the conscious mind, and that the unconscious level of awareness is characterized by the absence of the influence of biases. Further, he declared that '\textit{the unconscious typically involves a more objective and less distorted awareness of reality than the conscious}' (Havens, 2005).

\vspace{5mm}
\noindent \textit{Noise} 

\noindent Noise, or spontaneous firing, which is a known property of neurons, may cause longer jumps between conscious thoughts, especially when we lower our focus level. For the unconscious brain, it is reasonable to believe that spontaneous firing may be an even richer source to divergent thinking, since the unconscious is likely to be less affected by the focus level and the biases of the conscious mind. The important functions of noise in brain functioning, including for divergent thought and creativity, was thoroughly discussed by Rolls and Deco (2010).

\vspace{5mm}
\noindent \textit{Increased speed} 

\noindent Experiments with rodents have shown that the speed of the replay of memories during sleep increases multiple times. It appears that the unconscious thought processes may be accelerated in time during so-called sharp-wave ripples going on in the hippocampus (Liu et al,,2019). These are high frequent bursts of waves travelling through the neural network of the hippocampus and cortical regions. Ivry (1997) discusses the role of the cerebellum as a timing device which may control the speed of the replay. This small, but complex brain region is known to play an important role in controlling and coordinating bodily motor skills, but recent research points to the possibility that the cerebellum also plays an important role in creative processes (Ito, 1997; Vandervert, 2003). 

\vspace{5mm}
\noindent \textit{Reshuffling and randomization} 

\noindent  A recent study by Stella et al. (2019) on rats indicates that thoughts and memories are not only replayed during sleep, but sequences of memories may be randomized during replay. Liu et al. (2019) also found that humans unconsciously replay, reshuffle and reorganize experiences. This randomized replay may be an important source to unconscious divergent thinking where new combinations of thoughts and ideas are tested.

We might suggest that the brain conducts an open and objective dialogue, or even a multilogue, with itself beneath the surface of consciousness.  Thus, your unconscious brain acts as a highly sophisticated, multivariate statistical processor capable of generating a plethora of candidate thoughts and ideas by recombining and reshuffling learned sequences at high speed. The primary limitation is that much of this dialogue occurs beyond our conscious awareness, in the realm of the unconscious mind. Nevertheless, from time to time, this potent statistical mechanism brings forth an insight or a eureka moment into the forefront of our attention. However, for this to happen, the unconscious must not only present candidate ideas, but also evaluate their usefulness to some degree before any insight emerges into the mind’s eye. It is at this juncture that our biases, internal models, and structures play a crucial role.

The properties of stochastic processes discussed are significantly applicable to developing virtual association processes for Artificial Creativity (AC). Drawing inspiration from human divergent thinking, both conscious (serial) and unconscious (parallel), could enable the rapid and efficient generation of ideas.

Regarding the hypothesis from the Dartmouth proposal, which suggests that human creativity could be emulated \textit{in silico}, this might be considered the simpler aspect. It could also be argued that, through sheer computational power, a computer can explore a far broader space of idea combinations than the human brain. However, as will be illustrated later, most new ideas generated through the recombination of old ones could be largely futile without some form of guiding principle. McCarthy highlighted human intuition as a potential guide, and his insight into intuition likely wasn't far off. The divergent thinking characteristic of the human mind can be described as stochastic, associative processes that generally conform to the constraints (or biases) of a learned causal world model. Therefore, we might conclude that the process of divergent thinking is directed by our biases, strongly in the focused, conscious and weakly in the defocused, unconscious mind. Consequently, the human associative process could be a more effective generator of useful ideas than a computer’s exhaustive approach. 

\subsubsection{The importance of biases}

Cognitive bias is a well established term in psychology, and is an area of research trying to explain our tendencies to think, decide and act in ways that, in some objective sense, seems suboptimal. A wide range of cognitive biases have been defined, and predominantly, cognitive biases are given a flavor of being some kind of flaw in human cognition. 

Bias is also a central concept in statistics, where it expresses how the expectation of an estimator for some population parameter deviates from the true parameter value. Unbiasedness is therefore a quite common criteria in the search for good estimators, but there are many others. For instance, within Bayesian inference, uniform minimal risk is a more common criteria. Bayesian estimators therefore tend to be biased, but may have other benefits, like reduced variance or reduced overall risk. Unbiasedness is furthermore a property relative to some assumed true model, and this makes it quite context dependent. An estimator which is unbiased for a parameter under a given model may very well be biased under a wider more general model. 

Bias has increasingly become a part of the layman's vocabulary in recent years, particularly as decisions made by AI or outputs from generative AI, involving texts or images, have been shown to discriminate against certain groups. Statistically, this is recognized, especially in classification models trained on unbalanced data. Consequently, efforts are being made to mitigate this kind of bias through data balancing, aiming to adjust the training data to more accurately reflect a desired balance. However, such data filtering may result in models that act unbiased in one context but biased in another. Therefore, the pursuit of unbiasedness, while often desired, is far from straightforward and can be highly dependent on cultural factors. Paradoxically, in the quest to achieve AI with human-level intelligence and creativity, striving for unbiasedness seems contradictory when the essence of human intelligence and creativity may indeed be rooted in our biases. We will discuss why biases are crucial for creative exploration and learning.

When we use the term ‘bias’ in this paper, we have a quite wide interpretation in mind, representing some cognitive structure, either inherent or learned. Occasionally also other terms from psychology, neuroscience or statistics could be used quite synonymously, like: 'schema' (Jean Piaget), ‘reference frame’ (Milton Erickson), ‘belief’ or ‘prior’ (Bayesian statistics), or 'world model' (common AI terminology). Common to all is that they represent some subjective structure of knowledge or belief that we already possess. It might be anything from knowing how to discern apples from pears, to more abstract opinions about the existence of free will or God.

In the preceding section, we delved into the properties of the stochastic process of divergent thinking, which represents the initial stage of the creative process where potential novelties are proposed. This section shifts focus to how our biases influence divergent thinking by imposing constraints, as well as how we leverage biases to assess the utility of a divergent thought. Questions such as "Does it work?" and "Is it meaningful?" are crucial. Considering the potential of Artificial Creativity (AC), the aspects of human creativity discussed here will likely pose a more significant challenge to artificially implementing creativity than merely generating novel idea candidates. Nonetheless, it's essential to explore the role of biases in human creativity to understand whether it's possible to replicate or simulate human creativity in computers.

But before we turn to the importance of biases in the creative process, we will focus a bit on how our biases seem to affect both our perception and our attention.  

\vspace{5mm} 
\noindent \textbf{Biases as attention filters} 

\noindent The human brain is both conservative and novelty seeking. It is well known from psychology that we are biased towards trying to confirm what we believe to be true, our opinions and prejudices. This is known as confirmation bias. However, this is, luckily, balanced by an urge to seek novelty. The salience hypothesis in psychology addresses the question of what guides our attention, and it states that our attention tends to seek novelties or things that 'stick out' from some background in the environment. Our senses are receiving a tremendous amount of information every second as we are awake, and somehow the brain has to filter out most of this information as it seeks some attention point. It makes sense to think that we should attend to things that change or differ from some background signal. 

Surely, this is a good thing, for instance, if we drive along the street and a cat suddenly jumps into the road. The cat, representing a sudden change or surprise, attracts our immediate attention and gives us a chance to hit the breaks. However, a bird flying by in front of the car is perhaps less likely to give the same reaction, unless you are an ornithologist, perhaps. This example indicates that our attention is not only drawn towards novelty or surprise. This was demonstrated in study by Henderson and Hayes (2017).  They showed in experiments that visual attention is, in fact, also drawn towards meaning, and not surprise or novelty alone. This deviates from the salience hypothesis, which has been the dominant view in later years. 

Human attention is thus guided by top-down intrinsic bias, an inner motivation, guided by meaning, interest, values or feelings. We might say that we are drawn towards surprises that are meaningful to us. Without this top-down evaluation and filtering of novelty, we would be swamped in all the unexplainable noise that surrounds us everywhere. Hence, we might say that we are born to be creative in the way we naturally seek novelties that work. (Nakajima et al., 2019) recently hypothesized how the Striatum (part of the Basal Ganglia) regulates and filters input data, letting through relevant and blocking irrelevant information from our senses. Prat (2022) points to the so-called cortico-striatal pathway and dopaminergic reinforcement learning as a potential mechanism describing how the striatum learns from prefrontal cortex how to sort information according to our desires.

Besides serving as filters for attention, biases may have an even more direct influence on perception, for instance, on what we see or hear. A theory of mind and on human cognition is the theory of predictive coding (e.g. Friston and Kiebel, 2009; Millidge et al., 2021) stating that we all, in our daily lives, learn about and adapt to the environment by making predictions based on internal, cognitive models. We sense the world through our beliefs, and they are either retained or adjusted according to prediction errors. These errors are the differences between what we observe and our prior expectations based on the intrinsic models of our surroundings. So, we may say that we have internal, cognitive models of how things around us are expected to be and how they are predicted to change. Hence, the intrinsic models, or biases, therefore also influence how we perceive the world.

Thus, biases and internal models of our surroundings are important mental structures, essential for survival and well-being. Throughout life we learn to categorize sensory inputs and ideas to predict future outcomes in a random world. We build or adapt opinions, about values, interests, culture and ethical standards, all being mental reference frames helping us make decisions and bring meaning to our lives. Thus, through learning we build order in chaos, a hierarchical structure of biases, of boxes, to put life into. Some boxes are wide and big, like ‘beauty’ or ‘symmetry’, others are narrow and perhaps contained in others, like ‘carrot’, and ‘vegetable’. Many boxes we learn or adopt from others, some we creatively discover ourselves, and some we are probably even born into life with. 

\vspace{5mm}
\noindent \textbf{Biases limiting divergent thinking} 

\noindent As discussed in the previous section, the stochastics of divergent thinking, the associative thought-process can statistically be compared to a non-stationary, time-inhomogeneous stochastic process, where the transition probabilities over the flexible state space of thought change over time and with context. It is very plausible to assume that the transition probabilities, defining which thoughts are more likely to be associated together, depend on our biases; our learned structures of interests, values, opinions, prejudices and so forth. Hence, if we have strong biases, they are also likely to put limits to the process of divergent thinking, making it less likely to generate atypical ideas. 

During our lifetime we are constantly expanding our level of knowledge, building order in chaos, a complex bias structure, which may become increasingly rigid and non-flexible. This may put limits to creativity and explain why children seem to be more open to new ideas and be more creative than elderly people. Havens (2005) summarizes the insightful thinking of the famous psychotherapist Milton Erickson about this paradox of learning limiting new learning this way:

\vspace{5mm}

\noindent\textit{At first the ordinary person’s mind is relatively unstructured, objective, flexible and open to new learnings. Over time, however, it naturally becomes increasingly rigid, biased, idiosyncratic, and unable to accept perceptions, learnings, or responses that cannot be accommodated by its previously adopted structure. \ldots Eventually the entire conscious awareness of the individual may become restrictively governed or dictated by the very structure that originally developed to allow an increased freedom of response.}

\vspace{5mm}

David Bohm (2004) describes this as self-sustained confusion that can arise when a person’s mental frames have become so rigid and structured that any divergent thoughts challenging this mental structure become conflicting and painful. 

Sometimes this conflict is an inner conflict, but sadly the conflict may also be induced from the environment. One of the most popular TED-talks is the hilarious talk by Sir Ken Robinson with the slightly provocative title: “Do schools kill creativity?” Robinson makes a convincing argument that we are all born creative, but school has the unfortunate effect of making us suppress this inborn skill in the way our educational system is dominated by supervision, how it rewards conformity and punishes divergent thinking. Being 'wrong' is not accepted, and the result seems to be to surrender to self-sustained confusion.

\vspace{5mm}
\noindent \textbf{Biases and creativity} 

\noindent We will now briefly point back to the two step procedure for data simulation using Markov Chain Monte Carlo methods, which was used to exemplify a simple statistical creative process (see section \ref{divergent}) Initially we focused on the relevance of step 1 (divergent thinking) as part of creative processes, but for a divergent thought to be accepted as creative it has to be evaluated with regard to its usefulness. Every new thought aspiring to be deemed creative must be evaluated with regard to its usefulness within some reference frame or bias structure. It is common to use the phrase thinking outside the box to describe the process of coming up with new and unconventional thoughts that may lead to creative ideas, artwork or scientific discoveries. However, even though divergent thinking may lead to candidate ideas outside the box, we still need a box in order to evaluate the candidate idea. Creativity is always to think inside \textit{some} box, because without a box (a bias, or a frame of reference) the usefulness of a novel thought cannot be evaluated.

Our biases and reference frames may statistically be seen as priors in a Bayesian belief update process. Given our biases and mental models about how the world is, we evaluate the likelihood of new observations. If new observations seem trustworthy, but in conflict with our beliefs, we may “choose” to change our beliefs. On the other hand, if biases are strong and data are ambiguous, we may stick to our beliefs. This balance between the likelihood $P(D|M)$ of new observations (data) $D$ under the assumed world model $M$, and the prior probability $P(M)$ of the world model, is expressed by Bayes rule:

\[ P(M | D) \propto  P(M)\cdot P(D|M)\]

Thus, in an ongoing learning process, the prior belief $P(M)$ is updated to form posterior belief, $P(M|D)$, in light of the new observations. The posterior belief later serves as a
new prior belief as part of continued learning. 

However, if Bayes’ rule was the only way our priors could change, we would not be very creative. The Bayes rule expresses a rather slow change of prior beliefs, and novelties would find little support when evaluated using such priors. 
A more radical way to be creative is therefore to actively replace the prior in Bayes rule, that is, to transform or reconstruct our biases. 

Hence, creativity is most about challenging our old boxes, our reference frames. A divergent thought may be rejected as being creative according to one reference frame, but can, in fact, be accepted within a wider or transformed reference frame. Brovold (2014) expressed this similarly (translated from original): 

\vspace{5mm}

\noindent \textit{To creatively create order is not to follow conventions, but to include new elements into an order, preferably by establishing new categories and sorting principles, and tying these together in a different rhythm, form and function.... New products, new technology and new needs and services require such a continuous willingness to challenge, change, improve and include new elements in a partly old order.}

\vspace{5mm}

\noindent Based on our arguments so far we now suggest the following essence of human creativity:
\textit{The true act of creativity is the dynamical restructuring of biases or reference frames to guide the search for and the evaluation of novelties}

\vspace{5mm}

\noindent Hence, an artefact or an idea, like an actual painting, a sculpture, a new music play or even a new scientific theory, is not creative \textit{in itself}, and neither is the craftsmanship or the deductive reasoning that produced the sculpture or the new theoretical lemmas. Such manifestations of the creative insight into something observable is more a matter of skillful production or deduction (Bohm’s fancy). It is the imaginative and new insight of a new order, a new concept, or a new reference frame that leads to this manifestation, which is the true leap of creativity. This view also encapsulates another capability often included in the definition of human intelligence, namely the ability to adapt to an environment through a dynamical restructuring of reference frames.

However, both insight and fancy are necessary stages in a creative process in the way they work interchangeably and stimulate one another. This is also well known from design thinking principles. On the other hand, one may question whether a person, or a computer (as we will return to later) is creative in case he/she/it is only involved in the deductive reasoning step of the process. 

In the following we will look closer into how biases are used and transformed in different types of creative processes. This will make it easier to discuss the potentials of artificial creativity later.
 
\vspace{5mm}
\noindent \textbf{Supervised learning} 

\noindent Supervised learning may serve as a baseline with which to compare creative processes. If, for instance, a kid or a student acquires all knowledge and reference frames from external sources, like from a parent or a teacher, in an entirely bottom-up type of learning process, we may refer to the learning process as being supervised. Note that we here regard \textit{all} sensory input as bottom-up flowing information, that is, both what we in statistic would call dependent (supervision data) as well as independent data (explanatory data). Bottom-up learning means that information, possibly from many input sources (multimodal input), flows from input perceptions from our senses to be integrated and to form knowledge, but in a rather non-creative way. Without reflection or critical thinking this is mostly a process of transferring (or reconstructing) order and structure provided by the supervisor. A bottom-up learner would tend to follow instructions and never raise a critical question to the knowledge structures presented. Hence, pure bottom-up learning is not sufficient for creativity. 

\vspace{5mm}
\noindent \textbf{Imagination} 

\noindent Imagination, on the other hand, is a requirement for all types of creativity, and especially for the imaginative fancy stage of a creative process. Neurologically it is a top-down process, where the top level biases are fed back toward the sensory cortices to either alter factual perception of the world, or to (re)play 100\% imaginative experiences (Koenig-Robert, R. and Pearson, 2021), like we do in dreams. In relation to AI this can be regarded as a generative process. If you close your eyes and decide to imagine the face of a close relative or friend, information flows from higher order structures defining the person, down to your visual cortex area to create a mental picture of the face of your acquaintance. Hence, imagination is reverse application of cognitive biases by a top-down information flow. Such efferent (top-down) information pathways may also be central for our ability to causally predict or imagine possible futures, to make plans or to perform imagined experiments, like Albert Einstein did in his 'Gedankenexperimenten', which is an important part of Bohm´s fancy. Neurological studies support this. For instance, Kounios and Beeman (2015) found through EEG studies that people, who through tests appeared to be more imaginative than others, had higher resting state activity in the visual cortex. 

Finally, we may here mention that the role of imagination as a top down \textit{predictive} activity is in consistency with \textit{predictive coding theory} discussed above (e.g. Friston and Kiebel, 2009; Millidge et al., 2021). We may regard imagination as a the conscious part of the entire top-down predictive activity of the brain.

\vspace{5mm}
\noindent \textbf{Ways of transforming biases} 

\noindent Eagleman and Brandt (2017) found that most creative artefacts or ideas can be categorized into three main categories of novelty related to how they are created, either by \textit{bending}, \textit{breaking} or \textit{blending}. 
These categories are similar to the three-way classification of creative activities by Boden, (2004) who defined them correspondingly to \textit{exploratory}, \textit{transformational} or \textit{combinational}. In the following we will take inspiration from this and use a similar categorizations as Eagleman and Brandt (2017) and Boden (2004), but our angle is slightly different. We will apply them not to the outcomes of the creative process, but rather use the categories to describe how biases are transformed: Either you bend them, or you blend them, or you break them. We believe that lifting the focus like this, from the creations to the process leading to the creations, is necessary to take into account how adaptation and generalization are important aspects of human creativity. 

\vspace{5mm}
\noindent \textit{Bending} 

\noindent Relatively novel and useful creations may come about as small adjustments to already accepted ideas, theories or structures. A small change to established ideas is typically easy to suggest and readily acknowledged. Probably, a large part of published research can be described as adjustments to previous work, making minimal contributions towards a more complete knowledge structure of a given field or within an accepted paradigm. 
In terms of Bohm’s insight and fancy, bending is primarily applying imaginative and rational fancy to bend already accepted bias structures and hence, the level of creativity is low, and the manifestations can be characterized as variations of a old themes. 

\vspace{5mm}
\noindent \textit{Blending} 

\noindent Blending is a good characterization of creativity that is a result of applying and mixing previously learned structures to new areas. In this way order is extended by recognizing that old familiar structures may be 'blended' into the new and unexplored variability. 

Blending serves as a potent instrument for learning in general, extending beyond the realm of creativity. In the context of supervised learning, metaphors have been employed for millennia to aid in grasping and internalizing new concepts. Ancient instances include the Biblical parables and Plato’s allegory of the cave. Similarly, blending can assist the brain in discerning new patterns as part of a self-supervised creative process.

\vspace{5mm}
\noindent \textit{Breaking} 

\noindent Breaking describes creative processes where a completely new order that 'breaks' with previous order is created, and not just by a blending or bending of previous order. The son of Pablo Picasso once said that his father had the habit of 'breaking everything', but only to rebuild it in a novel and creative way. This is a perfect example of a manifestation of this last type of creativity. 

Breaking is a kind of self-supervised revision, transformation or extension of existing structures, and is likely the most challenging type of creativity in two ways. Firstly, as humans we tend to prefer to confirm old structures (confirmation bias). Breaking established order is for most people experienced as a stressful experience since the sense of stability is reduced or even destroyed. The scientist or the artist who questions structure must endure the emotional distress of the increased disorder and chaos that occurs before a new order is found. David Bohm writes about the self-sustained confusion maintained by those who cannot bear the distress of breaking old reference frames. This confusion may be sustained even if the old, and perhaps dear, biases clearly violate experience and perceptions. Rather than going through the pain of breaking and rebuilding biases, a person may confuse her-/himself by pointing to increasingly improbable explanations and exceptions. This may be a problem both in society at large, but even so in science. Scientists who have devoted their entire career within a scientific theory or paradigm may emotionally cling to the old biases and structures, rather than accepting that the contradictions between theory and data indicate the need of breaking and rebuilding theories.
[Bohm, 2004:]... \textit{he must be able to learn something new, even if this means that the ideas and notions that are comfortable or dear to him may be overturned.}

However, if the pain and confusion is overcome, and new, and more general order is found, the reward may overshadow the preceding distress. Apparently, some highly creative people, like Pablo Picasso, seem to handle this better than others and may even be curiously attracted to such a chaotic state. Also in scientific research it seems like some people can endure and perhaps be more attracted to the unresolved mysteries than others. Dörfler et al. (2018) describes a state called negative capability that scientists need in order to cope in times of transforming theories and changes of paradigms: 

\vspace{5mm}

\noindent \textit{Beyond the engagement with reality (and thus data), the negative capability is also important for an achievement of comprehension when we accept that the reality does not play by the management textbooks, and that researchers inevitably have to face a lack of internal consistency in their emerging understanding. Sometimes inconsistencies will disappear during the research project, but often they can persist for years. Thus, researchers need to develop an ability to cope with such a situation – and they need a framework in which a less than complete internal consistency can be accepted.}

\vspace{5mm}

For this type of creativity, requiring breaking of old reference frames, one may wonder how subjective usefulness is judged in the moment of insight in the creator. It is quite apparent that too narrow biases of the old structure must be ignored in favor of wider biases or prior distributions of acceptance of candidate ideas, as we would put it in statistical terms. It is likely that highly creative people use high level and wide biases, such as the sensations of wholeness, beauty and harmony, as guidance for their openness to new ideas. David Bohm states  (Bohm, 2004) that highly creative people, be it scientists or artists alike, seek a sense of wholeness, symmetry, harmony or beauty in their work:

\vspace{5mm}

\noindent \textit{The new order leads eventually to the creation of new structures having the qualities of harmony and totality, and therefore the feeling of beauty.}
 
\vspace{5mm}

This may also explain why simplicity, beauty, symmetry and totality tend to be more common measures of scientific validity of theories in some areas of mathematics and physics where there is a lack of experimental data for falsification studies, like for instance in some parts of cosmology.

Secondly, if an old, established structure is broken and a new order is suggested, it is very likely that the novel idea is met with scepticism, rejection,  or even ridicule in the community that still lives under the old structure. A scientist suggesting a shift of paradigm, or an artist creating a completely novel way of expression, risks that the difference in structural bias is too large for acceptance of the suggested new ideas in the community. The burden of proof of usefulness of the new idea, increases with the distance from the established biases.

\subsection{A neuroscientific view on human creativity}

Now it is time to ask to what extent is the philosophical view of Bohm and the statistical representation above reflected in neuroscience? Numerous neurological studies have been conducted using brain scans (e.g. fMRI) during conscious creativity tests, and researchers have gained some knowledge as to which parts or networks of the brain that are active during various stages of creativity. Some studies have shown that conscious divergent thinking and creativity is positively correlated with increased neural activity in the so-called default mode network (DMN) in the brain (Andrews-Hanna, 2011; Beaty et al., 2014). This network is typically active when we do not attend strongly to the outer world, but instead are focusing on internal goals, memories or planning. It is also shown to be active when we are preoccupied doing cognitively non-demanding tasks with low amounts of sensory input, like routine work, walking, showering or other moments of serendipity. Studies have also shown that other parts of the brain, making up the so-called central executive network (CEN), is more active when we are attending cognitively or emotionally challenging tasks, like problem solving and decision making. 

In parallel to the view that the creative process is an iterative process switching back and forth between the stages of insight and fancy, a creative person must own
enough flexible cognitive control to be able to switch effectively between the DMN and the CEN networks (Zabelina and Robinson, 2010). Recent neuroimaging studies support
this, showing that flexible and dynamic interactions between DMN and CEN is key to creativity (Beaty et al., 2018). Apparently, also a third cognitive network, the so-called
salience network (SN), is important here in controlling the interplay between DMN and CEN.

Hence, the quest for finding the connection between creativity and brain activity has lately switched from focusing on brain regions to considerations of network connectivity and flexibility. In section \ref{divergent} we connected such flexible cognitive control to the ability to switch easily between short and long transitions in the random walk of associative thinking, effectively switching between focused and diffuse states of mind. Apparently this is not only a matter of switching step lengths, but also switching between networks.

So the interplay between the three networks, DMN, CEN and SN, seems to correlate with a (semi-)conscious form of the creative process, in line with the rational insight and rational fancy stages of David Bohm. When it comes to imaginative insight, the Eureka moment type of creativity, a fourth brain network, has gained increasing attention in the last couple of decades. This is the so-called cerebro-cerebellar pathways connecting the cerebral cortices with the cerebellum. 

The importance of the cerebellum in fine-tuning and optimizing body movements (motor control) has been known for a long time. It is theorized that the cerebellum is creating fluent and advanced motor control by using predictive models combining automized movements (basic motor constituents) and performing continuous adjustments based on prediction errors. However, the importance of this brain region in mental processes, has a much shorter recognition in science. The role of the cerebellum also in creative processes has in later years been well elaborated and explored by Vandervert (2020) who relates the manipulations of this brain region to creativity by blending of thoughts: 

\vspace{5mm}

\noindent \textit{In sum, the cerebellum appears to play a predominant role in the refinement and blending of virtually all repeated movements, thoughts, and emotions.}

\vspace{5mm}

Although the cerebellum contains more than 75\% of all neurons in the brain, its cognitive processes are hidden in the unconscious. It is likely that the tremendous unconscious source of creativity discussed in the stochastics of divergent thinking is partly due to the skillful manipulations in the cerebellum. Vandervert (2003) carry forward the work by Ito (1997) in describing how the cerebellum unconsciously is generalizing thoughts (as well as movements) using so-called inverse dynamic models:

\vspace{5mm}

\noindent \textit{The inverse dynamics model helps explain how generalizations can be formed outside a person’s conscious awareness. This is a major reason that intuition may seem to leap out of 'nowhere'.} 

\vspace{5mm}

The imaginative insights of Bohm may thus be the outcome of these cerebellar processes, and we hypothesize that, in parallel to the conscious and rational insight discussed above, the cerebellum is supporting, and partly replacing, prefrontal cortex regions of the CEN and the SN in a creative communication with the DMN. The detachment of regions of the lateral prefrontal cortex in these kind of creative processes finds support in research by Limb and Braun (2008). They studied the stage of 'flow' in stages of improvisation among improvising jazz artists using fMRI scans. As they entered the flow state they registered reduced activity in the lateral prefrontal cortex, a part of SN and CEN (Limb and Braun, 2008; Magsamen and Ross, 2023). They also registered increased occurence of alpha brain waves, a characteristic of meditative and relaxed states with increased introspective focus and activity in DMN.

In this way conscious and serial manipulation of thoughts may hypothetically be replaced by unconscious, parallel manipulations controlled by the cerebellum.  During the flexible switching of the creative process, the CEN feeds the cerebellum with intentions, interests (biases) and problems to be solved. Further, the DMN provides a rich variety of reshuffled thoughts, memories and ideas, which the cerebellum recombines to fit intrinsic goal-oriented models. This is occasionally fed back to CEN to create imaginative and conscious insights, the eureka moments. This is in line with Vandervert (2020).

Just like for motor skills, like riding a bike or making the perfect golf swing, the cerebellum has automated primarily consciously controlled functions into unconscious routines. A question is how and when the imaginative insights are delivered to the cerebral cortex as conscious experiences of new ideas and orders. Obviously the news-content must be sufficiently important and relevant to a given context and problem to alert the conscious mind. We may imagine that sudden insight occurs when the unconsciousness finds a good posterior fit of (potentially reshuffled) new experiences into bias structures that we find particularly interesting. This may also partly explain why a-ha moments tend to occur during so-called incubation periods of serendipitous activity shortly after focusing hard on a given problem for some time. The focus period may simply increase the probabilistic value of the prior structures we find particularly interesting or promising for a given problem. This comes in addition to the fact that subsequent defocusing itself may help the signal reach the surface of consciousness easier, simply because the general attention level is lowered and competing focus points are few and weak (weak priors).
 
\subsection{An invariance perspective on human creativity}

We have discussed how creativity can be understood as the discovery of new order or structure by transforming and reshaping biases, typically by bending, blending or breaking the old mental frames. Since the semi-conscious or unconscious processes seem to be such a rich source of creativity, there appears to be favorable conditions for creative processes when the mind is unfocused. We have touched upon several potential factors in this paper that can partly explain this powerful property of the unconscious, like parallel processing, reshuffling of association chains, increased processing speed, and weakened biases. All these factors may together generate a favorable cognitive state for creativity; a state where the brain can exploit the fractal properties of biases due to invariance.

Fractal geometry is a well studied topic within an area of mathematics called chaos theory, and a fractal system is characterized by patterns that repeat or are self-similar at different scales. This part of mathematics is perhaps most famous from the Mandelbrot set (Brooks and Matelski, 1978; Mandelbrot 1980) as depicted in Figure \ref{fig:Mandelbrot_sets} (left). From the figure we can see swirls repeat at different scales as we dive into the picture. 
Several researchers have discussed fractal properties also in the structure of the human brain, for instance, in the way that the branching of the neural networks repeats at different scales. However, also in function the brain seems to be fractal, and this may explain how insight occurs, both consciously and unconsciously. The human brain is very good at finding similarities and familiar patterns everywhere. For instance, when we look up at the sky we may suddenly imagine the shape of a rabbit in a big cloud. Perhaps we even have to imagine the rabbit up-side down or with abnormally large ears or legs, but still we see it. The human brain is an extremely effective at pattern recognition, and it can easily bend, blend or break-and-rebuild structures by relocating, re-scaling and/or rotating old, familiar patterns. 

Hence, these patterns may be regarded as fractals which can be applied at different scales and in different locations and rotations. During this imaginative process scales seem to be non-important. In statistical terms we may use the term invariance to describe such a condition where location, scales or rotations don’t matter. It may be hypothesized that when we are in a semi-conscious or unconscious state of mind, the brain enters an invariant and fractal mode where new order is more easily generated through bending, blending or breaking. Even time may become invariant in the realm of the unconscious mind, just remember from Section \ref{divergent} how the speed of association processes are increased dramatically (shrinking time) and elements of thought chains are reshuffled (breaking chronology).

Liu et al. (2019), who found that the brain appears to be unconsciously reshuffling order of sequences in an attempt to fit new experiences into existing orders and structures, write:

\vspace{5mm}

\noindent \textit{Generalization of learned structures to new experiences may be facilitated by representing structural information in a format that is independent from its sensory consequences. \ldots Keeping structural knowledge independent of particular sensory objects is a form of factorization. Factorization (also called ‘disentangling’; Bengio et al., 2013; Higgins et al., 2017a) means that different coding units (e.g., neurons) represent different factors of variation in the world.} 

\vspace{5mm}

Factorization, disentangling and invariance are here all terms describing an objective and somewhat unstructured state that may be envisioned in the unconscious mind, wherein the cerebellum is free to make goal-directed fractal compositions.

This state of mind where order and structures are broken and transformed, resembles the phase transitions of matter in physics. For instance, when ice melts and forms liquid water, the tight order and structure of ice is broken, and a new structure is formed where water molecules move more freely. Recently a group of mathematicians (Duminil-Copin, et al., 2020) have come close to prove that so-called conformal invariance is a necessary property of phase transitions, the critical state between two phases of matter. Conformal invariance is a more extensive invariance which comprises all the three other invariances mentioned above; location (translational) invariance, scale invariance and rotation invariance. In our context it may be hypothesized that a state of conformal invariance also is a characteristic of the unconscious mind, and that the cognitive flexibility of insight and fancy can be compared to the phase transitions of matter. The creative process fluctuates between insight and fancy like some physical matter moves in and out of a critical state. From the highly invariant unconscious mind, new structure may crystallize as novel and useful ideas are popping into consciousness.

After this discussion on the stochastics of human creativity, we are now ready to look closer into the prospects of artificial creativity in the next section.

\section{On artificial creativity}

The year 2023 will perhaps be remembered as the year when artificial intelligence (AI) really became common property. In the course of the year so-called generative AI made it to the headlines and into our daily lives. Creating images, texts, music, programming code and even short movies from text prompts became mainstream and even habitual in short time. These applications are based on the latest advances in AI, like the transformer architecture (Vaswani et al., 2017), Large Language Models (LLMs) and diffusion models.

Generative AI separates from the more classical AI, (both symbolic AI and feed forward neural nets) which we may call inferential AI, in a way which we may readily relate to the above discussion on human cognition and creativity. Just as the information flow in the human brain can be separated into two main streams, the top-down (efferent) stream and the bottom-up (afferent) stream, we can regard generative AI as top-down prediction, and inferential AI as bottom-up inference or learning. From the AC perspective it is therefore the generative AI algorithms that has evoked most attention and emotional reaction due to the fact that these algorithms are able to generate output that may be perceived as creative.

Since the onset of the Second Industrial Revolution circa 1870, through the transition to the digital society around 1970, and until the last decade’s strong rise in artificial intelligence, the fear that the machines will take our jobs away has been there. And many tasks have also disappeared through automation of the most routine jobs in this period. However, throughout these stages of industrial and technological evolution, there have always been human attributes believed to be irreplaceable by machines—creativity being a prime example. This innate capacity for creativity has, to some extent, served as humanity's refuge against the methodical operations of machines and algorithms.

The strong emotions shown as a response to the latest developments in generative AI may reflect a belief that this haven is no longer safe, that we are now facing a new creative dimension in AI. In that case, this would be a radical step forward for AI. Could it be that creativity no longer is exclusive to humans? Headlines like CNN’s 'AI won an art contest, and artists are furious' are examples that show that this is the impression out there, and that this has the potential to change the creative professions. The Australian musician Nick Cave put words to many artists' concerns about generative AI replacing human creativity in his letter 'Why strive', which was read by the British comedian Stephen Fry at the Letters Live event in London in November 2023. Also the world champion in the boardgame Go, Lee Sedol, expressed that AI must have made a new leap as he was defeated by AlphaGo (Silver et al., 2017) in March 2016:

\vspace{5mm}

\noindent \textit{I thought AlphaGo was based on probability calculation and that it was merely a machine. But when I saw this move, I changed my mind. Surely, AlphaGo is creative}

\vspace{5mm}

However, as we’ve seen above, the human creative process involves more than the mere handcrafting or generation of a novel output, like an idea, a product, a song lyric, a new tune or even a Go move. For instance, it seems to involve an un- or self-supervised discovery of new order and structure, and the ability to evaluate usefulness of generated output. Will the algorithm stand up to the test in that respect? To try to answer this question and to evaluate the current status for artificial creativity (AC), we have to dig a little deeper into how contemporary generative AI works.

The basic principle behind many of the generative AI algorithms may shortly be described as an inferential part involving the mapping (projection) of observations, or part of observations/text (tokens) into a lower dimensional space through a compression process, often referred to as embedding. From a classical statistical point of view it is analogous to a latent subspace representation of data, like in principal component analysis. This can be regarded as the inferential (encoding) part of the model fitting process. The subsequent decoding step (prediction, data generation) will in various ways, and depending on the application, predict or generate novel, combined values (new instances
akin the input data) based on the nearest neighbour principle in the compressed latent space. The methods depend heavily on large amounts of labelled input training data, and predictions are based on correlations or distances in the latent space. 

Although the latent embedding space is heavily compressed compared to the dimensionality of the space spanned by the input data, the sub-space, or manifold spanned by the projected data is still large. LLMs and diffusion models therefore also have inbuilt attention mechanisms (Vaswani, et al.,2017), which based on the context of the surrounding text or the input prompt, is able to focus the data generation to seemingly more relevant parts of the manifold.   

This very short and heuristic description of the principles of generative AI serves the purpose as a back curtain for the following discussion on the creativity of AI. It clarifies that novelty is generated mainly from correlation/closeness principles in a latent space, and there are attention mechanisms that can guide the output generation to increase its relevance to some outer condition. 

It should be mentioned here that the game changer of AI for playing boardgames, the AlphaZero program of Deepmind (Silver et al., 2018), is based on other principles than this, namely a combination of reinforcement learning and tree search. Basically, the reinforcement learning part of AlphaZero consists of two deep neural nets, a so-called 'policy network' which chooses the next move and a 'value network' which predicts the winner. Since Alpha Zero seemingly showed both intuitive (as claimed by many for the precursor AlphaGo) and creative behaviour, we will also include Alpha Zero as we next look at some recent examples of AI that are relevant for a further discussion on AC.

\subsection{Examples of contemporary AI}

These examples convey a snap-shot of the current status in a field with a tremendous speed of progress. Hence, by the time this paper has completed the review process, these examples will probably already be yesterday's news. However, the principles for evaluating AC, which is presented here, is hopefully useful also for new advances, and the reader is encouraged to apply these ideas on new advances in AI.

\subsubsection{Example 1. Large Language Models }

Large Language Models (LLMs), like those part of ChatGPT (OpenAI), Bard (Google), Gemini (Google Deepmind) and Bing (Microsoft), are basically advanced 'next token' (or lets say 'word') predictors. Given a set of previous words or a textual context, the algorithms predicts the next words or token in a sentence among the closest neighbours in the latent embedding space. A user defined parameter called the 'temperature' controls the randomness of the output. If the temperature is set to a high value, the algorithm is more open to choose less probable words on the list of closest neighbours. This increases the variability of the output, and often this is even referred to as a more 'creative' setting of the LLM. The question is whether texts written by LLMs indeed can be creative?

With reference to section \ref{divergent} and the stochastics of human divergent thinking, the text generation algorithm may similarly be regarded as a Markov process, where a next token is proposed through a random walk process given the previous history of generated text. The initial prompt provided by the user may be regarded as the \textit{primer} for the process and the temperature as the \textit{focus level}. 

Statistically speaking this process resembles the Metropolis-Hastings algorithm, where an initial proposal step generates a proposed next token of the sentence. The next step would be an evaluation step where the algorithm judges whether the proposed token should be accepted as useful or not. This parallels the two steps of a creative process as described thoroughly in this paper. 

As we've concluded that novelty is generated through the random walk Markov process, what can be said about the evaluation of usefulness? As soon as the next token is proposed, there is no inbuilt mechanism for accepting or rejecting the proposal with regard to its likelihood of serving a useful purpose, fitting a context or expressing a desired meaning. The probability of accepting the proposal as the next predicted token is equal to 1. This means that an underlying assumption is that every new token is indeed useful and sampled from a desired bias structure. However, the well known hallucination problem of LLMs should be proof that this assumption is not fulfilled. Hence, by the description of creative processes given above, the LLMs are lacking the necessary capability to judge whether its own creation is indeed useful or not.

It should be added here that attention mechanisms, that may take a specific context into account, and fine-tuning of models, seem to increase the quality of the proposed
texts as judged by human users. Statistically this can be described as replacing the full prior with a conditional prior, conditional on the given context. However, this is only increasing the likelihood of making proposals within the bias structure learned by the model, not serving as an evaluation mechanism. This should increases the probability
that a generated text indeed is a meaningful and relevant response to the given prompt, but only unintentionally so, and with regard to thinking outside the box, it is difficult to see that this focusing of the model should make it more creative. The evolution of LLMs during the later years, with increased size and improved attention mechanism, first seemed to show an increase in relevance of the output. This may be due to the fact that the models were taking larger chunks of the generated text as a given context for the prediction of the next token. However, there might be a sweet spot here for how much the context should be accounted for. A highly specified context may lead to a very narrow conditional distribution for next tokens prediction and reduced randomness. This
may be the reason why the latest LLMs seem to regurgitate longer sequences from the training data. One consequence of this is the lawsuit in Dec 2023 by New York Times
against OpenAI for copyright infringement based on multiple examples of long parts of articles ’re-generated’ by ChatGPT. Some have been quite astonished that the LLMs
seem to ’remember’ longer parts of the training data like this, but the explanation for this non-creative regurgitation is likely more the extremely narrow conditional prior used for next token prediction leading to a very high probability for the very same token that
also is the next token in the most relevant training text. Hence, increased attention and focus works contrary to creativity.

We should not forget here that there are algorithms for evaluating the quality of longer generated texts (more than the next token), especially in the context of translation between languages. However, in such cases it is relatively easy to define usefulness (matching translation), but in these cases creativity should not be of interest. 

Finally, we need to mention that there indeed is a way to evaluate the usefulness and quality of generated texts, namely through human feedback. So-called reinforcement learning through human feedback (RLHF) is an established quality assessment and model improvement method. Of course, this human intervention just pinpoints the lack of automatic evaluation schemes in LLMs. 

Compared to human creativity LLMs are at most comparable to the 'bending' type where new variations (mixes) of old ideas within an existing bias structure are generated. Any usefulness is unintentional and merely a product of thinking inside the box given by the training data. The model is incapable of transforming its bias structure through blending or breaking.

\subsubsection{Example 2. Diffusion models }

\begin{figure}
    \centering
    \includegraphics[width=1\linewidth]{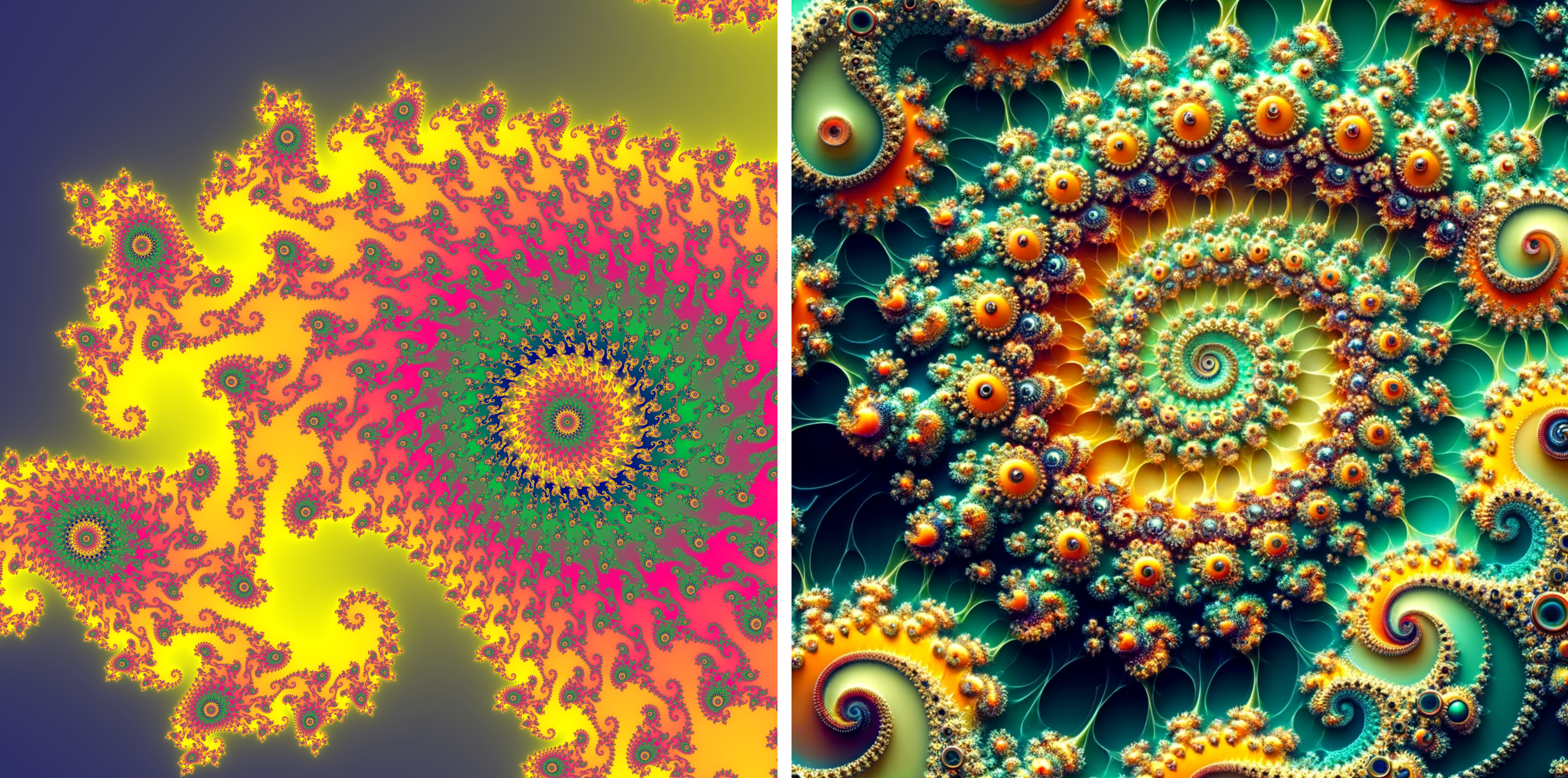}
    \caption{Mandelbrot sets from the 'seahorse area'. Left: Exact (as generated by \href{https://mandel.gart.nz/?Re=-0.7436448564&Im=0.1318267539&iters=1347&zoom=61744873185&colourmap=0&maprotation=0&axes=0&smooth=1}{mandel.gart.nz}). Right:(as generated by DALL-E 2, OpenAI)}
    \label{fig:Mandelbrot_sets}
\end{figure}

The ability to generate images from text prompts is perhaps the kind of generative AI that most people today find as ’creative’. This is perhaps mainly due to the fact that this is activity accepted as creative in the \textit{loose} sense, as discussed above. The quality of the images generated by the diffusion models is impressive and the algorithms are steadily improved to match the content of the prompts. An example is the Mandelbrot Set in Figure \ref{fig:Mandelbrot_sets} (right). This picture was generated via the prompt: \textit{Generate an image of the Mandelbrot Set around the point c=-0.74364 + 0.131825i}. Without going into the mathematics of the Mandelbrot set generation, we can add here that the exact Mandelbrot Set in the leftmost figure in Figure \ref{fig:Mandelbrot_sets} we generated around the same parameter \textit{c} as provided in the prompt to DALL-E 2 here. Both figures address a region called the Seahorse Valley in the Mandelbrot Set.

At first glance the images seem quite similar with repeating swirls at different scales, however, a closer look at the AI generated image reveals that the swirls are different and not self-repeating, and some are more like circles, or nobs. Although the image might be characterized as novel, and even beautiful, it does not obey the mathematical fractal structure as shown in the exact Mandelbrot figure to the left. The reason lies in the way that the diffusion algorithm creates a blended image based on similar images in the training data, and the algorithm has no knowledge of the deeper and common causal structure behind the images it has been trained on. The result is a novel blending of outputs, but it would not serve the purpose in a mathematical context. For the reader, however, the image may still be evaluated as creative in the way it is pleasing or beautiful to look at, but this is in the eye of the beholder, not the machine. 

This example illustrates that the diffusion models is capable to generate candidate values as a mixture of closest neighbours on the relevant part of the manifold trained by the model. Apparently, this is a clever way to generate novelty that 'might work'. Obviously images generated in this way contain recognizable elements, and often answering to the given text prompt, but what can be said about the evaluation step in the creative process? To what extent is the product evaluated as useful?

The image diffusion models are generating image proposals according to a bias structure in the sense that the latent manifold of the encoder model reflects the biases of the (human) creators of the input images. Hence, the algorithm might generate useful and creative outputs as a result of operating within the mix of priors established by the original artists and illustrators that created the training samples. However, similar to the LLMs this is a quite unintentional way to achieve novelty that works. This is so because also these algorithms have no inherent evaluator of its own creation. There are no mechanisms that evaluates the candidate image with regard to beauty, symmetry, causal plausibility, or obeying mathematical rules (as the Mandelbrot example). It is still the human user of the generative AI tool who is left to 'cherry pick' from the proposed images what might 'work' for a certain purpose, and thereby judging its creative value. 

\subsubsection{Example 3. Reinforcement learning}

Machine learning is typically divided into three main categories, unsupervised learning, supervised learning and reinforcement learning. In the context of creativity, reinforcement learning is particularly interesting in the way that an agent is learning new order and structure guided by feedback from its environment during training. As we've been arguing in this paper, creativity requires the ability to apply, discover or create order through bending, blending or breaking of biases that successively can be used to evaluate the usefulness of proposed novelties. As a successful example of the use of reinforcement learning, we will discuss the creative properties of the AlphaZero algorithm from DeepMind (Silver et al., 2018).  

The boardgame Go was unquestionably a huge challenge for AI, and with $10^{170}$ possible board configurations, it goes without saying that any supervised learning algorithm approaching this as a closed-system of finite possibilities comes short. It therefore attracted much attention when DeepMind developed their AlphaGo algorithm (Silver et al., 2017), which in 2016 defeated the world champion Lee Sedol 4-1 in a machine versus man combat across 5 games. The AlphaGo algorithm and its successors, like AlphaZero, combines a tree search algorithm and two neural nets for reinforcement learning. With the two stages of creative processes in mind, we find that AlphaZero fits seemingly well into the pattern of seeking 'novelties that work'. The algorithm has two main components, a 'policy network' that combined with a Monte Carlo tree search, serves as a proposal algorithm for the next move, and a 'value network', which serves as an evaluation function to judge the usefulness (predicting the expected winner) of the chosen move. This model can also be described as a Markov decision process (MDP) (Puterman, 2014).

As the 'Zero' in AlphaZero represents, the algorithm is initiated without any historical knowledge of previous human played games (which AlphaGo in fact had). Hence, in the beginning there is no 'bias' which could be used to guide or evaluate actions. The algorithm is only familiar with its environment (the board) and the basic rules, and it teaches itself from scratch to master the game through self-play. Through millions of simulated games played against itself, each leading to a win or a loss, AlphaZero built experience through trial-and-error. It is from this historical base of observations the algorithm is capable to build a prior to guide the algorithm only to consider a limited set of most promising moves, and to evaluate the value for a set of future actions.
This fits into the way we've described the role of biases earlier in this paper: to guide the proposal distribution and to evaluate the usefulness. So, can we then claim that the AlphaZero system is indeed creative? 

It is an intriguing fact that a human master of the game only considers about some 100 possibilities for each move, whereas AlphaZero considers about 100-1000 times more moves, and still they perform at about the same level. Furthermore, the AlphaZero has through self play experienced a much larger historical base of board configuration than any human can accomplish in his/her lifetime. So how can humans still play at the same level? 

The answer to this may lie in the role and use of biases. The value function of the reinforcement algorithm is (once the model is trained) a static prediction model of the winner of the game, given the proposed actions. This means that model has no opportunity to dynamically transform or reconstruct its prior model, and hence rather limited capability to evaluate the usefulness of radical moves outside its domain of historical observations (even though this domain is huge). 

We've earlier described the essence of human creativity as '...the dynamical restructuring of biases or reference frames to guide the search for and the evaluation of novelties', and in light of this it is hard to conclude that AlphaZero is creative at a human level. We have earlier discussed how this dynamical restructuring of biases can come about in humans through processes exploiting the state of conformal invariance. This process of invariant pattern recognition may also be an important difference between man and machine here. A human may for instance recognize similarities in broad patterns between different board configurations and identify promising moves in the game by the type of creativity referred to as 'blending'. This may happen both consciously and unconsciously, and often we may hear players use intuition as an explanation for their moves in retrospect.

\subsubsection{Example 4. AI for Scientific research}

Scientific progress is regarded as a highly creative activity, and the pushing the boundaries of knowledge often requires new perspectives and occasionally even leads to shifts of paradigms where 'world models' are broken down and rebuilt. At least until now, computers, statistics, machine learning and AI have only been valuable \textit{tools} for researchers. However, Google DeepMind have recently released several systems, like FunSearch and AlphaGeometry, that allegedly are able to push the boundaries of scientific knowledge. Since the systems are built around similar workhorses, we will use FunSearch as the example here.

FunSearch (Romera-Paredes et al. (2024)) is a new advanced AI system for mathematical discoveries, and the system was applied on various mathematical challenges to illustrate its potential. The approach uses LLMs to transform an initial pool of program codes into a mutated program that could happen to be a better solver for a given mathematical problem. In the next step the candidate is scored using an evaluation function that measures the quality of the candidate solution and makes sure they work as intended. The best programs are fed back into the pool of mutable programs. By iterating between the mutation step (making proposals) and the evaluation step millions of times, FunSearch was shown to find novel solutions, for instance, for the 'cap set' problem. Thus, FunSearch seems to meet the criteria in the 'novelties that work' sense of creativity. But is it really creative? On their web-pages DeepMind gives this impression in their description:

\vspace{5mm}

\noindent \textit{'FunSearch is an iterative procedure; at each iteration, the system selects some programs from the current pool of programs, which are fed to an LLM. The LLM creatively builds upon these, and generates new programs, which are automatically evaluated.} 

\vspace{5mm}

However, a seed code that generates the initial pool of programs, and more importantly, the evaluation function (problem description) used in the scoring of candidates, must be provided by the user. The evaluator is a static function (unless changed by the user), and the program is therefore not able to automatically reconstruct or change its bias structure, as we've discussed earlier. Therefore, FunSearch is best described as an evolutionary algorithm which generates candidates within a fixed bias structure defined by the user and that of the LLM. In light of the above discussion on the creativity of LLMs, we must conclude that the only creative force in FunSearch is still the human behind the wheels. FunSearch is unquestionably a clever creation by DeepMind as a tool for scientific discovery through the way it generates 'novelties that might work', but it seems to be oversold when it comes to its creative capabilities.

\section{Discussion}

All the examples of AI that have been mentioned here are extremely powerful and advanced AI algorithms that accomplish tasks that were pure science fiction only a decade ago. The development within reinforcement learning and generative AI has been mind blowing in many ways. It is not surprising that the quality of texts, images, music an so on, generated by these systems leads us to think that AI has become creative and that AGI is near. 

However, for AI to reach human level intelligence, it must also reach human level creativity. In this paper we argue that a key to human creativity lies in our cognitive biases, because every new thought aspiring to be deemed creative must be evaluated with regard to its usefulness within some reference frame or bias structure. Furthermore, creativity may require a restructuring or transformation of old bias structures in order to see a new purpose or meaning. This transformation is a bottom-up/top-down dynamical and iterative cognitive process by which we change the way we perceive the world as a response to our perceptions and actions. Bohm described this creative process as a continuous cycle of imaginative and rational stages of insight and fancy. We've seen through several examples of state-of-the-art AI algorithms that they to some extent fulfill the requirement of creating something new, but they do not stand up to the test when it comes to the ability to evaluate the usefulness of their creations in a dynamical and transformative manner.

Reinforcement learning algorithms come close to this kind of dynamical process in the training phase where it could be argued that the bias structure is subject to bending, but this plasticity is not present in a pre-trained model. The ability to use blending or breaking of priors appears still to be only a human capability even if blending seem to be within the repertoire of generative AI for proposing new candidate results. Wiggins (2019) concludes similarly, at least in the case of transformational creativity which corresponds to 'breaking': 

\vspace{5mm}
\noindent \textit{The application of transformational creativity means that the agent is a self-organising, self-evaluating system, and this is far beyond the scope of traditional AI search.} 

\vspace{5mm}
\noindent One possible reason to this is the lack of a functional world model in modern machine learning models. 
Generative AI algorithms are still, thus, unable to understand causal relationships to the extent that it can be used to carry out planning or thought experiments. This is important to be able to assess how a new idea or product will work if they were to be realized. Similarly, no algorithm is yet able to assess whether text, theories, formulas or references that it generated for scientific purposes, is actually in line with previous knowledge or basic axioms within a field unless specific and fixed evaluation functions are provided. 

In the context of AGI this lack of a world model in current deep learning and generative AI has thoroughly been addressed by Marcus and Davis (2019). They point to hybrid
models combining both deep neural networks and so-called symbolic AI as a promising way forward in this respect. Later LeCun (2022) proposed a system architecture inspired
by the brain, including a world model module and a perception-action loop, as a possible path towards human level AI, or autonomous intelligence, as he named it. Interestingly, autonomy in transforming own reference frames (world model) would also be required by
a system with truly creative capabilities, as discussed in this paper. Ma et al. (2022) also suggested a closed-loop system where a bottom-up and a top-down information flow ensures a certain parsimony and self-consistency. This kind of closed-loop encoder/decoder systems would probably be needed for an artificial agent to interact with and learn continuously and creatively from its environment. In this way bottom-up perception (or imagined perception) meets top-down world model predictions possibly leading to transformed biases through Bayesian belief updates. What we just described here fits perfectly into the predictive coding theory (e.g. Friston and Kiebel, 2009; Millidge et al., 2021). We therefore believe that the most promising path towards Artificial Creativity is within this framework of this ’theory of mind’ which dates all the way back to Helmholz (1867).

\section{Conclusion}
The conclusion that we draw from this is that we have yet to see algorithms that exhibit human level artificial creativity. Algorithms can in no way replace human creative competence as of today.
However, the new AI algorithms can be considered useful tools that can help increase the exploration of new territory and come up with suggestions within the creativity category of bending. For more ground-breaking creativity, within the categories blending or breaking, however, AI seems to come short in every respect, still.  Whether or not the final creation, be it art, music, innovation or research, can be considered creative, will for now have to be left entirely to the human being who uses these tools.

\section*{Acknowledgements}
This article is partly based on first author's Metacognition blog \url{https://blogg.nmbu.no/solvesabo/}.

\section{References}

Andrews-Hanna, J.R., 2012. The brain’s default network and its adaptive role in internal mentation. The Neuroscientist, 18(3), pp.251-270.

Beaty, R.E., Benedek, M., Wilkins, R.W., Jauk, E., Fink, A., Silvia, P.J., Hodges, D.A., Koschutnig, K. and Neubauer, A.C., 2014. Creativity and the default network: A functional connectivity analysis of the creative brain at rest. Neuropsychologia, 64, pp.92-98.

Beaty, R.E., Kenett, Y.N., Christensen, A.P., Rosenberg, M.D., Benedek, M., Chen, Q., Fink, A., Qiu, J., Kwapil, T.R., Kane, M.J. and Silvia, P.J., 2018. Robust prediction of individual creative ability from brain functional connectivity. Proceedings of the National Academy of Sciences, 115(5), pp.1087-1092.

Bengio, Y., Mesnil, G., Dauphin, Y. and Rifai, S., 2013. Better mixing via deep representations. In International conference on machine learning (pp. 552-560). PMLR.

Boden, M.A., 2004. The creative mind: Myths and mechanisms. Psychology Press.

Bohm, D., 2004. On creativity (Vol. 13). Psychology Press.

Brooks, R. and Matelski, J.P., 1981. The dynamics of 2-generator subgroups of PSL (2, C). In Riemann surfaces and related topics: Proceedings of the 1978 Stony Brook Conference (Vol. 1). Princeton, New Jersey: Princeton University Press.

Brovold, H., 2014. Invarians drøftet i et nevropsykologisk perspektiv med spesiell referanse til realfaglig kognisjon.‘Fire veier inn i matematikken’. NTNU-trykk. (In Norwegian). \url{https://ntnuopen.ntnu.no/ntnu-xmlui/handle/11250/271186}.

Dörfler, V., Stierand, M. and Chia, R., 2018. Intellectual quietness: our struggles with researching creativity as a process. In BAM 2018: 32nd Annual Conference of the British Academy of Management.

Duminil-Copin, H., Kajetan Kozlowski, K., Krachun, D., Manolescu, I., and  Oulamara, M., 2020. Rotational invariance in critical planar lattice models. \newline   https://doi.org/10.48550/arXiv.2012.11672 

Eagleman, D. and Brandt, A., 2017. The runaway species: How human creativity remakes the world. Catapult.

Friston, K. and Kiebel, S., 2009. Predictive coding under the free-energy principle. Philosophical transactions of the Royal Society B: Biological sciences, 364(1521), pp.1211-1221.

Gilks, W.R., Richardson, S. and Spiegelhalter, D. eds., 1995. Markov chain Monte Carlo in practice. CRC press.

Hastings, W.K., 1970. Monte Carlo sampling methods using Markov chains and their applications, Biometrika, Volume 57, Issue 1, Pages 97–109

Havens R. A., 2005. The wisdom of Milton H. Erickson. Crown House Publishing.

Henderson, J.M. and Hayes, T.R., 2017. Meaning-based guidance of attention in scenes as revealed by meaning maps. Nature human behaviour, 1(10), pp.743-747.

Higgins, I., Pal, A., Rusu, A., Matthey, L., Burgess, C., Pritzel, A., Botvinick, M., Blundell, C. and Lerchner, A., 2017. Darla: Improving zero-shot transfer in reinforcement learning. In International Conference on Machine Learning (pp. 1480-1490). PMLR.

Hubert, K.F., Awa, K.N. and Zabelina, D.L., 2024. The current state of artificial intelligence generative language models is more creative than humans on divergent thinking tasks. Scientific Reports, 14(1), p.3440.

Ito, M., 1997. Cerebellar microcomplexes. In J. D. Schmahmann (Ed.), The cerebellum and cognition
(pp. 475–487). New York: Academic Press.

Kahneman, D., 2011. Thinking, fast and slow. Allen Lane.

Koenig-Robert, R. and Pearson, J., 2021. Why do imagery and perception look and feel so different?. Philosophical Transactions of the Royal Society B, 376(1817), p.20190703.

Kounios, J. and Beeman, M., 2015. The Eureka factor: Creative insights and the brain. Random House.

LeCun, Y., 2022. A Path Towards Autonomous Machine Intelligence Version 0.9. 2, 2022-06-27.

Limb, C.J. and Braun, A.R., 2008. Neural substrates of spontaneous musical performance: An fMRI study of jazz improvisation. PLoS one, 3(2), p.e1679.

Liu, Y., Dolan, R.J., Kurth-Nelson, Z. and Behrens, T.E., 2019. Human replay spontaneously reorganizes experience. Cell, 178(3), pp.640-652.

Ma, Y., Tsao, D. and Shum, H.Y., 2022. On the principles of parsimony and self-consistency for the emergence of intelligence. Frontiers of Information Technology \& Electronic Engineering, 23(9), pp.1298-1323.

Magsamen, S. and Ross, I., 2023. Your Brain on Art: How the Arts Transform Us. Random House.

Mandelbrot, B.B., 1980. Fractal aspects of the iteration of z $\rightarrow$ $\Lambda$z (1‐z) for complex $\Lambda$ and z. Annals of the New York Academy of Sciences, 357(1), pp.249-259.

Marcus, G. and Davis, E., 2019. Rebooting AI: Building artificial intelligence we can trust. Vintage.

Millidge, B., Seth, A. and Buckley, C.L., 2021. Predictive coding: a theoretical and experimental review. arXiv preprint arXiv:2107.12979.

Nakajima, M., Schmitt, L.I. and Halassa, M.M., 2019. Prefrontal cortex regulates sensory filtering through a basal ganglia-to-thalamus pathway. Neuron, 103(3), pp.445-458.

Prat, C., 2022. The neuroscience of you: how every brain is different and how to understand yours. Penguin.

Puterman, M.L., 2014. Markov decision processes: discrete stochastic dynamic programming. John Wiley \& Sons.

Ritchie, G., 2019. The evaluation of creative systems. Computational creativity: The philosophy and engineering of autonomously creative systems, pp.159-194. 

Rolls, E.T. and Deco, G., 2010. The noisy brain: stochastic dynamics as a principle of brain function.

Romera-Paredes, B., Barekatain, M., Novikov, A., Balog, M., Kumar, M.P., Dupont, E., Ruiz, F.J., Ellenberg, J.S., Wang, P., Fawzi, O. and Kohli, P., 2023. Mathematical discoveries from program search with large language models. Nature, pp.1-3.

Silver, D., Schrittwieser, J., Simonyan, K., Antonoglou, I., Huang, A., Guez, A., Hubert, T., Baker, L., Lai, M., Bolton, A. and Chen, Y., 2017. Mastering the game of go without human knowledge. nature, 550(7676), pp.354-359.

Silver, D., Hubert, T., Schrittwieser, J., Antonoglou, I., Lai, M., Guez, A., Lanctot, M., Sifre, L., Kumaran, D., Graepel, T. and Lillicrap, T., 2018. A general reinforcement learning algorithm that masters chess, shogi, and Go through self-play. Science, 362(6419), pp.1140-1144.

Spearman, C., 1923. The nature of" intelligence" and the principles of cognition. Macmillan.

Sternberg, R. J., 1985. Beyond IQ: A Triarchic Theory of Intelligence. Cambridge University Press.

Vandervert, L., 2003. How working memory and cognitive modeling functions of the cerebellum contribute to discoveries in mathematics. New Ideas in Psychology, 21(2), pp.159-175.

Vandervert, L., 2020. How the cerebellum and cerebral cortex collaborate to compose fractal patterns underlying transpersonal experience. International Journal of Transpersonal Studies, 39(1), p.19.

Vaswani, A., Shazeer, N., Parmar, N., Uszkoreit, J., Jones, L., Gomez, A.N., Kaiser, Ł. and Polosukhin, I., 2017. Attention is all you need. Advances in neural information processing systems, 30.

Von Helmholtz, H., 1867. Handbuch der physiologischen Optik: mit 213 in den Text eingedruckten Holzschnitten und 11 Tafeln (Vol. 9). Voss.

Wiggins, G. A., (2019). A framework for the description, analysis and comparison of
creative systems. In T. Veale \& F. A. Cardoso (Eds.), Computational creativity: The philosophy and engineering of autonomously creative systems (pp. 21–48). Springer.

Zabelina, D.L. and Robinson, M.D., 2010. Creativity as flexible cognitive control. Psychology of Aesthetics, Creativity, and the Arts, 4(3), p.136.

\end{document}